\documentclass{article}

\usepackage{times}
\usepackage{graphicx} 
\usepackage{subfigure} 

\usepackage{natbib}

\usepackage{algorithm}
\usepackage{algorithmic}

\usepackage{hyperref}

\usepackage{icml2016}

\usepackage{amsmath}
\usepackage{amssymb}
\usepackage{wrapfig}
\usepackage{float}
\usepackage{multirow}
\usepackage{hyphenat}

\newcommand{\R}{\mathbb{R}}

\newcommand{\I}[1]{\operatorname{\mathbb{I}}{[#1]}}

\DeclareMathOperator*{\argmin}{argmin}
\DeclareMathOperator*{\maximize}{max\;}

\DeclareMathOperator*{\subjectto}{subject\ to\;}

\icmltitlerunning{A Hierarchical Spectral Method for Extreme Classification}

\begin{document} 

\twocolumn[
\icmltitle{A Hierarchical Spectral Method for Extreme Classification}


\icmlauthor{Paul Mineiro}{pmineiro@microsoft.com}
\icmlauthor{Nikos Karampatziakis}{nikosk@microsoft.com}
\icmladdress{Cloud and Information Services Laboratory, 1 Microsoft Way, Redmond, WA 98052}

\icmlkeywords{extreme learning, label decomposition, spectral methods}

\vskip 0.3in
]

\begin{abstract} 
Extreme classification problems are multiclass and multilabel
classification problems where the number of outputs is so large that
straightforward strategies are neither statistically nor computationally
viable. One strategy for dealing with the computational burden is via
a tree decomposition of the output space.  While this typically leads
to training and inference that scales sublinearly with the number of
outputs, it also results in reduced statistical performance. In this work,
we identify two shortcomings of tree decomposition methods, and describe
two heuristic mitigations.  We compose these with an eigenvalue technique
for constructing the tree.  The end result is a computationally efficient
algorithm that provides good statistical performance on several extreme
data sets.
\end{abstract} 

\section{Introduction}

Classification applications with large numbers of possible outputs,
aka extreme classification problems, are becoming increasingly
common, e.g., language modeling, document classification, and
image tagging.  Furthermore, reductions of structured prediction
\citep{daume2014efficient} and structured contextual bandit
\citep{chang2015learning} problems to classification can induce
large numbers of possible outputs.  Consequently, a robust extreme
classification primitive would enable new approaches to problems such
as ranking, recommendation, or interactive learning.

When the number of possible outputs is small, the structure of the output
space can be ignored.  Although this presumably sacrifices computational
and statistical efficiency, this can be overcome via brute-force,
i.e., additional time and space complexity for training and inference,
and additional sample complexity for training.  For a large number of
outputs, however, this is insufficient: in practice it is necessary to
tackle issues such as inference time and model size while retaining good
generalization, and these goals are facilitated by output structure. For
some applications the algorithm designer can posit an approximately
correct output structure.  Extreme classification focuses on applications
where the output structure is unknown and needs to be inferred.

In this work we revisit tree based decomposition algorithms for extreme
classification. Although this class of algorithms enjoy favorable
inference (and training) time, their accuracy is typically worse than
direct approaches such as a flat softmax. We proceed to identify two
aspects of tree based algorithms that negatively affect their statistical
performance. First, training data is decimated as we go deeper down
the tree, so there is not enough data to learn flexible models as the
tree grows deeper.  Second, trees are not robust to mistakes due to
incorrect routing at the internal nodes of the tree.  We then propose
techniques that reduce the effect of these shortcomings. Finally we
present an efficient eigenvalue-based algorithm that learns a tree
which maps an example to a (small) set of candidate outputs;
this tree is coupled with an underlying generic classification strategy
for a complete solution.

\section{Background and Rationale}

\subsection{Notation}

We denote vectors by lowercase letters, e.g., $x$, $y$; matrices by
uppercase letters, e.g. $W$, $Z$; and sets by calligraphic uppercase
letters, e.g., $\mathcal{A}$, $\mathcal{B}$.  The input dimension is
denoted by $d$, the output dimension by $c$ and the number of examples
by $n$. For multiclass problems $y$ is a one hot (row) vector, i.e.,~a
vertex of the $(c-1)$ unit simplex; while for multilabel problems $y$ is
a binary vector, i.e.,~a vertex of the unit $c$-cube.  The power set of
labels is denoted by $\mathcal{P}(c)$. The indicator function is denoted
as $\I{p}$ and is $1$ if $p$ is true and $0$ otherwise.

\subsection{Tree-Based Decomposition Algorithms}

Many existing algorithms that enjoy inference (and sometimes training)
times that scale sublinearly with the number of classes are based
on partitioning the output space with a tree structure. Early works
(e.g.\ \cite{morin2005hierarchical}) treated the tree as given
while more recent papers \cite{mnih2009scalable, bengio2010label,
prabhu2014fastxml, choromanska2014logarithmic} learn both the tree
and the classifier. However, if a flat softmax classifier is 
feasible\footnote{today, circa 100,000 unique labels is the practical limit.}
it will typically obtain better test accuracy~\cite{rifkin2004defense}.
There are several reasons why label decomposition via trees can
underperform. In the sequel, we describe two of them. Our proposed
solutions are described in detail in the next section.

\textbf{Data decimation} refers to the reduced amounts of data that
reach each individual node in the tree. Label trees are typically trained
in a top-down fashion, with the root having access to all the data and
learning a router that sends examples to the left or right child. Each
child sees only the examples that get routed to it and learns how to
route this subset to its children. This continues recursively until the
leaves of the tree so each leaf receives a disjoint set of the training
examples. Typically an output space with $c$ labels leads to a tree
with $O(\log c)$ levels and a constant number of candidate classes at
each leaf. Therefore each leaf classifier only has access to $O(n/c)$
examples. Given this, it is not surprising that some of the label tree
algorithms employ severely limited classifiers at the leaves, such
as static predictors that ignore all information about the incoming
example. Instead, we propose to treat the set of classification 
tasks at each leaf node as a multi-task 
learning problem~\cite{caruana1997multitask} and leverage
techniques from that literature, e.g., sharing classifier parameters
across leaves.

\textbf{Routing brittleness} refers to the sensitivity of the
prediction to mistakes in routing. If the leaves form a partition of
the classes then a single mistake anywhere along the path from the
root to a leaf will cause an example to be misclassified.  Therefore,
a typical countermeasure is to have each class in multiple leaves
in the hope that the classifier can recover from some mistakes during
routing. However, at training time we have additional information about
which examples are likely to have problems with routing via the margin
associated with the routing decision. We should expect that examples
from the true distribution similar to low margin examples could route
differently. While we cannot access the true distribution, we can assign
an example fractionally to both the left and right branches using the
margin.  This will make the training algorithm more robust to those parts
of the input space that fall near the decision boundaries of the routers.

\section{Algorithm}
\subsection{Design}

The basic idea is to identify a (small) set of candidate labels for the
example, and then invoke a base classifier only on the candidate set.
Intuitively, the function which determines the set of candidate labels
should have high recall, whereas the base classifier should have high
precision when restricted to the candidate set, such that the composition
is accurate.  We can use any base classifier with our technique, so the
focus of this paper is learning the candidate label set function.

\begin{algorithm}[H]
\caption{Predict($F, H, x$)}
\begin{algorithmic}[1]
\STATE $\mathcal{C} \leftarrow F (x)$\hfill \COMMENT{$F: \mathcal{X} \to \mathcal{P}(c)}$
\STATE Return $H (x, \mathcal{C})$\hfill \COMMENT{$H: \mathcal{X} \times \mathcal{P}(c) \to \mathcal{Y}$}
\end{algorithmic}
\label{alg:cset}
\end{algorithm}

Our architecture is succinctly described in Algorithm~\ref{alg:cset}.
We utilize a function $F$ which maps an example to a small set of
candidate labels, along with a more generic classifier $H$ whose possible
outputs are limited to those produced by $F$.  For this construction
to provide time complexity reduction, the cost of invoking $F$ should
be small, the set $\mathcal{C}$ returned by $F$ should typically be
small, and $H$ should have complexity which is independent of $c$
given $\mathcal{C}$. For example, it could depend only upon
the size of the candidate set rather than the total number of labels.

To limit the cost of invoking $F$ we will use a tree-based decomposition
with limited depth.  To limit the cost of invoking $H$, we place
an upper bound on the size of the set returned by $F$. This creates
two hyperparameters, the tree depth and the leaf node class budget;
increasing either will typically improve accuracy at the cost of
additional computation. Section~\ref{sec:experiments} provides some 
guidance on selecting these hyperparameters.

Ensuring $H$ has computational complexity dependent only upon the
size of the candidate set, rather than the total number of labels, is
idiosyncratic to the underlying classifier used.  We describe how to
modify the two models used in our experimental section.  For multiclass
logistic regression, we only compute and normalize
the predictions over the candidate set.  For multilabel per-class
independent logistic regression, we only compute
the values in the candidate set.  These modifications can also be applied
to the final layers of deep architectures.

\subsection{Stagewise Learning}

Our ultimate research goal is an online algorithm for jointly learning
$F$ and $H$, and some of our design choices have been motivated by
plausibility of adaptation to the online setting.  In this work,
however, we restrict our focus to batch stagewise learning of $F$ and $H$, 
in which $F$ is first constructed and then $H$ is optimized given $F$.

To learn the function $F$ which determines the set of candidate labels, we
will recursively construct a tree whose nodes route the examples to 
their children. Each router is found by solving an eigenvalue problem which
attempts to purify the label distribution in the induced subproblems.
The candidate sets for each leaf node are the labels with
highest empirical frequency. 
While the relationship between the ideal objective (recall) and our eigenvalue
objective is indirect, our approach is computationally scalable and 
empirically achieves high recall.

\subsection{Learning a Tree Node}

Clearly, if the true label(s) are not contained within the candidate set
$\mathcal{C}$ than the overall procedure must commit an error. This
suggests that learning $F$ should be done by maximizing recall-at-$k$.
Finding the tree that maximizes recall-at-$k$ is a hard combinatorial 
problem and instead we follow a top-down procedure which optimizes a
different objective at each node in the tree.
Nonetheless, we compute recall as a diagnostic for $F$ in our experiments.

For multiclass problems, \cite{choromanska2014logarithmic} show that
arbitrarily low error rates can be achieved by a tree-style decomposition
under a weak hypothesis assumption.  The weak hypothesis assumption states
it is always possible to find a hypothesis which achieves a minimum level
of purity while limited to a maximum level of balance.  Purity refers to
the larger of the fraction of each class' examples which routes to the
left or the right, and is ideally close to 1; whereas balance refers to
the larger of the fraction of all examples which route to the left or the
right, and is ideally close to 0.5.  Inspired by this, our approach is
to optimize a related purity objective subject to an approximate balance
constraint, in order to achieve a label filtering with high recall.

We temporarily restrict our attention to multiclass problems.  For a
linear routing node given instances $(x, y)$ ranging over $\R^d \times \{
0, 1\}^c$, a direct implementation of \cite{choromanska2014logarithmic}
with a strict balance constraint would find a weight vector $w \in \R^d$
such that \[
\begin{aligned}
\maximize_{\substack{w \in \R^d \\\|w\|=1}} & 
\sum_c \left| \mathbb{E} \left[ \I{w^\top x > 0} y_c - \I{w^\top x \leq 0} y_c\right] \right|
\\
\subjectto & 
\mathbb{E}\left[ \I{w^\top x > 0} - \I{w^\top x \leq 0}\right] = 0 
\end{aligned}
\] We replace the balance constraint with $w^\top \mathbb{E}[x] = 0$.
While this generates splits that are close to balanced,
we find improved results via computing the empirical median $b =
\mathrm{median}(w^\top x)$ and using $\I{w^\top x > b}$ for routing.

We replace the purity objective with
\begin{equation}\label{eq:purity}
\sum_c \left( \mathbb{E} \left[ w^\top x {y_c} \right] \right)^2 
=w^\top \bigl( 
\sum_c \mathbb{E} \left[ x {y_c} \right] \mathbb{E} \left[x {y_c} \right]^\top 
\bigr) w,
\end{equation}
which is an eigenvector problem on the sum of the frequency-weighted
class-conditional means.  Note the substitution
\begin{equation*}
\begin{split}
&\sum_c \left| \mathbb{E} \left[ \I{w^\top x > 0} y_c - \I{w^\top x \leq 0}y_c\right] \right| \\
&\;\;\longrightarrow \sum_c \left( \mathbb{E} \left[ w^\top x y_c \right] \right)^2
\end{split}
\end{equation*}
replaces the original purity objective (which tries to maximize the
per-class absolute difference in examples routing left vs. right) with
a proxy (which tries to maximize the distance between the projected
class-conditional mean and the origin). While \eqref{eq:purity}
admits a fast solution, the objective is now sensitive
to the magnitude of the examples. This can 
be mitigated by preprocessing the data so that the norms of the
examples are similar. In Section~\ref{sec:experiments} most datasets 
are processed such that all examples have unit norm.

To generalize to multilabel, we condition on the entire label vector for
each example: on a data set of features $X \in \R^{n \times d}$ and
labels $Y \in \{ 0,1 \}^{n \times c}$, and a linear predictor $\hat X =
Y (Y^\top Y)^{-1} Y^\top X$ of $X$ given $Y$, equation~\eqref{eq:purity}
is equivalent to $w^\top (\hat X^\top \hat X) w$.

Ultimately we arrive at the constrained eigenvalue problem
\begin{equation}
\maximize_{w:\|w\|=1, 1^\top X w = 0} w^\top (\hat X^\top \hat X) w. \label{eqn:eigenprob}
\end{equation}
Except for the balance constraint, equation~\eqref{eqn:eigenprob}
corresponds to one variant of orthonormal partial least squares.  This is
sensible given the bilinear factor model interpretation of PLS: we seek
a direction in feature space which captures the maximum variance in the
label space.  Thus while equation~\eqref{eqn:eigenprob} was motivated
by analysis for the multiclass case, it is also plausible for multilabel
problems.

Equation~\eqref{eqn:eigenprob} is a top eigenvalue problem which can
be efficiently solved, e.g., using power iteration or randomized PCA
\citep{halko2011finding}. We can incorporate the linear constraint
by projecting out $1^\top X$ during power iteration. 

The dependence upon the number of classes $c$ is via the cost of
multiplying with the hat matrix $Y (Y^\top Y)^{-1} Y^\top$.  The 
hat matrix is idempotent which implies $\hat X^\top \hat X = X^\top \hat
X$ and eliminates one application of the hat matrix.  For multiclass
the remaining application can be done in $O(n)$, i.e., constant time
per example.  For multilabel we exploit that multiplication with the
hat matrix is equivalent to solving a least squares problem, i.e.,
\begin{equation}
(Y^\top Y)^{-1} Y^\top X z = \argmin_v \| Y v - X z \|_2^2.
\label{eqn:leastsquares}
\end{equation}
Because in our multilabel experiments $Y$ is very sparse, we use a small
number of iterations of diagonally preconditioned conjugate gradient
which is sufficient for good results.  Each iteration is only $O(n s)$,
where $s$ is the average number of nonzero labels per example.

\subsection{Recursive Tree Construction}

To learn an entire tree, we combine the previous procedure for learning
a single node with a specification of how to recursively create child
problems.

Once a node has been constructed, a straightforward way to induce
subproblems is to route the data set according to the routing function
at that node. Unfortunately, however, typically a constant fraction of 
examples are very close to the boundary.  Moreover, the
routers along each path are diverse which implies that most examples
have multiple, almost independent, ``chances'' of being routed with
a low margin for some router in the tree. From the point of view of generalization this is particularly worrisome: each leaf is defined 
by an intersection of surfaces (half-spaces for linear routers) and,
for deep enough trees, most examples are near at least one of these
surfaces. Moreover, each node completely ignores the examples from 
``neighboring'' tree nodes that could be used for, say, smoothing 
out the estimates of the most frequent classes in that node.

We mitigate this problem by defining a routing probability at each 
node and using fractional routing during tree building and randomized routing during training of the underlying classifiers,
which we describe in detail in the sequel. We utilize fractional and randomized routing only during training. At inference time, we use deterministic routing.

Fractional routing means that when we construct $F$, we propagate expected example counts to each
child node, optimize an importance-weighted version of
equation~\eqref{eqn:eigenprob}, and utilize aggregates of expected counts to determine which classes are in the candidate set at a leaf.

The routing probability at each node is heuristically defined by assuming that a test example
$\tilde{x}$ is sampled from a Gaussian centered at a training example
$x$.  Let $w$ and $b$ be the routing vector and bias for a particular
node.  Under the above assumption we have
\[
p(w^\top \tilde{x} > b) = p(w^\top x + \sigma w^\top z > b ) 
= \Phi\left(\frac{w^\top x - b}{\sigma}\right)
\]
where $z$ is a vector of iid standard Gaussians and $\Phi$ is the CDF of
a standard Gaussian.  The first equality is by definition of $\tilde{x}$
while the second is by rearranging terms and using that $-w^\top z$ has
a standard Gaussian distribution (since $||w||=1$ c.f.\
\eqref{eqn:eigenprob}). We heuristically set the standard deviation
$\sigma=\lambda/m$ where $m$ is the sum of the importance weights of all
examples routed to the node, and $\lambda$ is the eigenvalue from
equation~\eqref{eqn:eigenprob}.

We define the probability of an example along a path in the tree to be the product of its routing probabilities over the nodes in the path.  
These probabilities also allow for terminating the recursion prior to
hitting the maximum allowed tree depth, because the expected counts can
be used to estimate recall-at-$k$.  Thus, we can specify an additional
hyperparameter $\phi$ which is the acceptable recall-at-$k$ at which to
terminate recursion.

The fractional example counts are equivalent to importance weights in the
eigenvalue problem of equation \eqref{eqn:eigenprob}, and in particular
for multilabel classification we solve an importance-weighted version
of the least squares problem in equation \eqref{eqn:leastsquares}.

Once the tree is constructed we have a function from example to leaf node.
We then associate a set of classes with each leaf node in order to
define the function from example to set of classes.  The set of labels
associated with a leaf node is defined as the $k$ most frequent labels in
the training set which route to that node (weighted by the probability
of them reaching the node), where $k$ is the leaf-node class budget
(a hyperparameter which limits training and inference time complexity).

\subsection{Classifier Training}

Once the tree has been constructed, we optimize the classifier $H$ using
stochastic gradient descent.  We use randomized routing during this stage
to sample a path through the tree each time an example is encountered
during optimization.  The sampled path associates a candidate set of labels with each
example according to the leaf node terminating the path.  We then modify
the training (and inference) procedure to honor the restriction of the
model to the candidate set.  For multilabel this is straightforward as
we use an independent logistic link for each class.  For multiclass we
use softmax link, in which case we compute output pre-activations only
for the candidate set, and then normalize only over the candidate set.
In both cases time complexity is independent of the total number of
classes given the cardinality of the candidate set.

Note the same underlying model $H$ is being employed at every leaf node.
This is motivated both by the desire to mitigate space complexity and
the need to limit model complexity given the decimation of data that
occurs at the leaves.  However, we augment the original features with a
categorical variable indicating the leaf node.  This allows us to treat
the problem of training the underlying classifier as a multi-task learning
problem, where the node id indicates the task.  At one extreme, ignoring
the node id means the same classifier is used at every tree node (albeit
leveraging a different candidate set of labels).  At the other extreme,
each task can be considered completely separate, i.e., each leaf node
has a distinct classifier.  The latter option is typically not viable
for large data sets either computationally (as the space complexity is
quite high) or statistically (as data decimation at the leaves would
force an excessively simple classifier).  Intermediate possibilities
from the multi-task learning literature include multiplexing via a hash
kernel \citep{weinberger2009feature}, positing a low-rank structure on
the parameters across tasks \citep{evgeniou2007multi}, and leveraging
group-sparsity \citep{chen2011integrating}.  For our experiments we find
empirically that only modest customization of the underlying classifier
per-node is statistically viable; for more details see section
\ref{sec:experiments}.

\section{Related Work}

Tree based decompositions are popular in the extreme learning
literature due to their inherent scaling properties. In early
work \cite{morin2005hierarchical} on language modeling the tree was
derived from a hand crafted hierarchy and given to the algorithm as an
input. More recent work has focused on learning the tree structure
as well \cite{mnih2009scalable, bengio2010label, prabhu2014fastxml,
choromanska2014logarithmic}.  Like our technique here, tree based
decompositions recursively partition the feature space in order to induce
easier subproblems.

Our randomized routing approach is similar to \cite{mnih2009scalable}
in which a training example is sent both left and right if it is within
$\epsilon$ (a hyperparameter) of the router's decision boundary.

Some techniques focus solely on improving inference time for a given
model.  In \cite{bengio2010label} spectral techniques are used to
define a tree based decomposition via the eigenvectors of a matrix
derived from the confusion matrix of a pretrained (multiclass) model.
The composition of a label filter with a base classifier was proposed by
\cite{weston2013label}, whose inference procedure is essentially the same
as our proposal.   During training, however, we learn the filter prior to
learning to underlying classifier rather than the converse, which allows
us to exploit the computational speedup of filtering and consequently
utilize a more computationally expensive classifier.  Indeed, the relative poor comparative performance
of \cite{weston2013label} to FastXML reported in \cite{prabhu2014fastxml}
is plausibly due to the use of a na{\"i}ve Bayes classifier to mitigate
computational constraints during training.

The idea of recursively solving eigenvalue problems to convert a
multiclass problem into a sequence of simpler classification problems 
was explored by~\cite{yildiz2005linear}.  In that work authors
heuristically searched over possible partitionings of the classes at each
node, which becomes increasingly infeasible as the number of classes
increases.  Furthermore, their technique does not apply to multilabel problems.

Randomized methods for efficient eigenvalue decomposition were
introduced by \cite{halko2011finding} and are useful to the practical
implementation of our approach.

Embedding based approaches are also popular in the extreme learning
literature.  These techniques seek a low-dimensional representation of
the features and/or labels which mitigate both computational and sample
complexity.  We compare experimentally with X1 \citep{bhatia2015locally},
an embedding method with state of the art performance on several public
extreme data sets.

Algorithm~\ref{alg:cset} is similar in spirit to the operation of modern
search engines, in which hand-crafted features such as BM25~\cite{robertson1999okapi} 
are used to filter the set of candidate documents prior to more expensive and higher 
precision re-ranking by another model.

\section{Experiments}\label{sec:experiments}

Software to reproduce these experiments is at
\url{https://anonymized}.

\begin{table}[h]
\centering
\caption{Data sets used in experiments.  $s$ is the average number of labels per example.}
\begin{tabular}{|c|c|c|c|c|c|c|c|} \hline
\multirow{2}{*}{Dataset} & \multirow{2}{*}{$s$} & \multirow{2}{*}{$n$} & \multirow{2}{*}{$d$} & \multirow{2}{*}{$c$} & Root Node \\
& & & & & Learn Time  \\ \hline 
Twitter & 1.27 & 25M & 1M & 264K & 49s \\ 
ALOI & 1 & 97K & 128 & 1K & 0.3s \\
ODP & 1 & 1.5M & 0.5M & 100K & 2.5s \\
LSHTC & 3.26 & 2.4M & 1.6M & 325K & 13s \\ \hline
\end{tabular}
\label{tab:datasets}
\end{table}

Table~\ref{tab:datasets} lists the datasets used in our experiments,
along with the time to learn the root node of the tree.  All times quoted
in the experimental section are for a Matlab implementation running on a
standard desktop, which has dual 3.2Ghz Xeon E5-1650 CPU and 48Gb of RAM.

In principle, several parallelization strategies are available.  First,
eigenvalue problems are inherently amenable to distributed implementation,
as the computational bottleneck is matrix-vector product.  Second, all
nodes at a particular depth of the tree can be learned in parallel, i.e.,
parallel running time for computing the entire tree is a function of the
depth given sufficient resources.  In practice for these experiments we
compute the entire tree sequentially on a single machine.

Many published algorithms do not scale to the datasets utilized in our
experiments section. Even among those that do, replication on these
datasets is challenging.  Therefore our baseline comparisons for these
experiments, while seemingly idiosyncratic, are the current published
state-of-the-art procedures for these datasets.  We also compare with
\cite{choromanska2014logarithmic} when possible, as our technique is
inspired by their analysis.

Regarding hyperparameters: for these experiments, we found that
increasing the number of candidates per leaf node $k$ always improved
results statistically, and therefore $k$ was set in practice by our own
notion of acceptable training time of the underlying classifier (note
tree construction time is independent of $k$).  However, increasing
the tree depth did not always improve results statistically, therefore,
while building the tree we monitored a hold-out set for recall in order
to determine the best depth.

\subsection{Twitter}

Unlike the other data sets in this section, there is no widely used
classification problem associated with this data set. Instead, we 
utilized this dataset to explore the label structure uncovered by the 
tree learning algorithm.
Twitter hashtags are convenient for this purpose, as they are numerous,
interpretable, and strongly related to the text of the containing tweet.
We took a 6 months sample of Twitter data from the beginning
of 2010, filtered for tweets containing a hashtag, and then filtered
out hashtags which did not occur at least 5 times in the sample.  This
resulted in 25 million tweets containing 264 thousand unique hashtags.
We used a 20-bit hashing kernel with unigrams and bigrams to generate a
feature representation for each tweet, and use a primal representation of
the Hellinger kernel, i.e., we normalized each tweet's token frequency
to sum to 1 and then took the square root of the token frequencies.
This ensures each tweet's features has unit Euclidean norm.


\begin{table}[h]
\centering
\caption{Selected nodes' most frequent classes (hashtags) for the tree learned from Twitter data.}
\begin{tabular}{|c|} \hline
Tags  \\ \hline
\parbox{0.9 \linewidth}{ \vspace{2pt} \centering \nohyphens{ 
\#nowplaying \#jobs \#ff \#fb \#tweetmyjobs \#news \#dc \#stl \#sf \#pdx \#1 \#raleigh \#austin \#sac \#nashville \#followfriday \#phoenix \#ny \#pittsburgh \#la
} } \\ \hline
\parbox{0.9 \linewidth}{ \vspace{2pt} \centering \nohyphens{ 
\#vouconfessarque \#nowplaying \#ff \#bbb \#bbb10 \#jobs \#douradofacts \#todoseriador \#fail \#cpartybr \#fato \#haiti \#maiorabracovirtual \#dourado \#fb \#livres2010 \#oremos \#coisasdetimido \#qualquergarota \#todoadolescente
} } \\ \hline
\parbox{0.9 \linewidth}{ \vspace{2pt} \centering \nohyphens{ 
\#nowplaying \#ff \#jobs \#retweetthisif \#bieberbemine \#happybirthdayjustin \#babyonitunes \#biebcrbemine \#justinbiebcr \#fb \#tweetmyjobs \#damnhowtrue \#followfriday \#biebcrgasm \#1 \#grindmebieber \#quote \#news \#retweetthis \#followmejp
} } \\ \hline
\parbox{0.9 \linewidth}{ \vspace{2pt} \centering \nohyphens{ 
\#jobs \#it \#nowplaying \#manager \#dev \#engineering \#ff \#java \#marketing \#php \#job \#net \#project \#developer \#hiring \#programmer \#engineer \#consultant \#customer \#flash
} } \\ \hline
\parbox{0.9 \linewidth}{ \vspace{2pt} \centering \nohyphens{ 
\#nowplaying \#ff \#jobs \#donttalktome \#retweetthis \#shooturself \#deleteyouraccount \#letsbehonest \#thisdateisover \#sheprobablyahoe \#retweetthisif \#howwouldyoufeel \#fb \#unwifeable \#urwack \#cantbemyvalentine \#tweetmyjobs \#fail \#imthetypeto \#1
} } \\ \hline
\end{tabular}
\label{tab:twitter}
\end{table}

Table~\ref{tab:twitter}
shows the most frequent classes (hashtags)
for selected nodes from a depth 12 tree.  The procedure is capable
of discovering clusters of hashtags that are related by functional,
regional, or semantic cohesion.  However, the most frequent hashtags
(e.g., \#nowplaying, \#ff) are essentially placed in every node.
We return to this issue in the discussion section.

\subsection{ALOI}

ALOI is a color image collection of one-thousand small objects
recorded for scientific purposes~\citep{geusebroek2005amsterdam}. The
number of classes in this data set does not qualify as extreme by
current standards, but we utilize it to facilitate comparison with the
underlying classifier alone.  For these experiments we will consider test
classification accuracy utilizing the same train-test split and features
from \cite{choromanska2014logarithmic}.  Specifically there is a fixed
train-test split of 90:10 for all experiments and the representation is
linear in 128 raw pixel values.  Unlike the other experiments, the ALOI
examples have variable Euclidean norm, but the
variation is modest: 95\% of the examples have norm between 
34 and 148.

\begin{table}[h]
\centering
\caption{ALOI purity results, comparing the eigenvalue technique to
random root nodes.  In all cases train balance is 50\% by construction.}
\begin{tabular}{|c|c|c|c|} \hline
Method & Equation~\eqref{eqn:eigenprob} & Average & Maximum\\ \hline
Train Purity & 86.7\% &  71.9\% & 84.8\% \\ \hline
Test Purity  & 86.9\% &  74.0\% & 85.6\% \\ \hline
Test Balance & 49.98  &  50.0\% & 49.96\% \\ \hline
\end{tabular}
\label{tab:purityaloi}
\end{table}



ALOI is a multiclass dataset so we can investigate the relationship
between the eigenvalue problem and the original multiclass purity
objective.  Table~\ref{tab:purityaloi} shows the results.  We computed
the training and test purity for the split induced by the root node. We
also sampled 10,000 random weight vectors uniformally distributed on
the hypersphere and computed the same quantities.  For this dataset,
the purity at the root node is better than that achieved by the maximum
over the random draws, indicating the eigenvalue procedure is achieving
an extreme quantile of the purity distribution, despite optimizing a
proxy. This advantage is maintained on the held-out test data.  In all
cases, perfect balance on the training set is achieved by construction.  
Test set deviation of balance is negligible.

\begin{table}[h]
\centering
\caption{ALOI results.  Averages are example-averages across the training set. }
\begin{tabular}{|c|c|c|} \hline 
Model                                          & Rank-50 LR    & \parbox{36pt}{\centering \vspace{1ex} Tree + Rank-50 LR \vspace{1ex}} \\ \hline
{\centering \vspace{1pt} Avg Depth}            & n/a           & 13.94 \\ \hline
{\centering \vspace{1pt} Avg Leaf Classifiers} & n/a           & 24.09 \\ \hline
{\centering \vspace{1pt} Test Tree Recall}     & n/a           & 96.5\% \\ \hline
{\centering \vspace{1pt} Test Accuracy}        & 91.12\%       & 91.03\% \\ \hline
{\centering \vspace{1pt} Inference Speed}      & 125,000 ex/s  & 41,000 ex/s\\ \hline
\end{tabular}
\label{tab:aloiresults}
\end{table}

Classification results are shown in Table~\ref{tab:aloiresults}.
The baseline is a rank-50 logistic regression, i.e., a single hidden
layer neural network with 50 hidden nodes and linear hidden activations,
trained via stochastic gradient descent.  The comparison is a tree
combined with rank-50 logistic regression, where the weights from input
to the 50-dimensional intermediate representation are shared across
all tree nodes but the output bias and the weights from intermediate
representation to output are node-specific.  We used a tree with maximum
depth 12, maximum classifiers per leaf of 25, and acceptable training
tree recall of 99.9\%.

Generalization is comparable between the two solutions, with the errors
induced by the tree partially compensated by the increased flexibility
of the underlying classifier due to multi-task training: in contrast, 
when using the same classifier at every leaf, training
accuracy decreases from 95.97\% to 95.43\% and test accuracy 
from 91.03\% to 89.53\%.

Empirically we find randomized routing is an effective regularizer.
If we use deterministic routing during tree construction, training tree
recall increases from 98.1\% to 100\% but test tree recall drops from
96.5\% to 93.3\%.  If we use randomized routing during tree construction
but deterministic routing while training the underlying classifier,
training accuracy increases from 95.97\% to 97.90\% but test accuracy
drops from 91.03\% to 79.34\%.

Inference times are dominated by constant factors, but note that the tree
solution does 38 hyperplane evaluations per example, while the baseline
does 1000.  On the larger datasets analogous differences can translate
into wall clock advantage if they overwhelm constant factors.

\cite{choromanska2014logarithmic} report 86.5\% test accuracy on this
data set.  However,
their procedure is an online learning procedure which is applicable in
scenarios where our stagewise learning procedure is not.

\subsection{ODP} 

The Open Directory Project \citep{ODP} is a public human-edited
directory of the web which was processed by \cite{bennett2009refined}
into a multiclass data set.  For these experiments we will consider test
classification error utilizing the same train-test split, features, and
labels from \cite{choromanska2014logarithmic}.  Specifically there is
a fixed train-test split of 2:1 for all experiments, the representation
of document is a bag of words, and the unique class assignment for each
document is the most specific category associated with the document.

We used a tree with maximum depth 14, maximum classifiers per
leaf of 4000, and acceptable training tree recall of 99.9\%.
The underlying classifier is a logistic regression, using a hashing
kernel \citep{weinberger2009feature} to $2^{15}$ dimensions to mitigate the space complexity.
After hashing we use a primal representation of the Hellinger kernel, i.e., 
we normalized each document's token frequencies to sum to 1 and then took 
the square root of the token frequencies. 

The node indicator variable is augmented to the feature representation
but not interacted, i.e., each node learns a distinct bias vector for
the logistic regression, but otherwise shares all parameters.

\begin{table}[h]
\centering
\caption{ODP results.  Averages are example-averages across the training set.}
\begin{tabular}{|c|c|c|} \hline 
Model                                          & Rank-300 LR         & Tree + LR  \\ \hline
{\centering \vspace{1pt} Avg Depth}            & n/a                 & 13.98 \\ \hline
{\centering \vspace{1pt} Avg Leaf Classifiers} & n/a                 & 3882 \\ \hline
{\centering \vspace{1pt} Test Tree Recall}     & n/a                 & 50.4\% \\ \hline
{\centering \vspace{1pt} Test Accuracy}        & 16.85\%             & 19.53\% \\ \hline
{\centering \vspace{1pt} Inference Speed}      & 1700 ex/s           & 230 ex/s \\ \hline
\end{tabular}
\label{tab:odpresults}
\end{table}

Results are shown in Table~\ref{tab:odpresults}.  The baseline is
a rank-300 logistic regression,  i.e., a single hidden layer neural
network with 300 hidden nodes and linear hidden activations, trained
via randomized techniques \citep{mineiro2015fast}, which is the current
state of the art for this data set.  Computationally, the baseline
is superior to the current method despite considering every class at
inference time, because it can fully exploit the hardware (vectorization,
cache locality, etc.).  Our label filtering method, while asymptotically
faster, cannot exploit all the features of the architecture it is
running on, both because the sequence of vector products associated
with routing in the tree is conditional on previous results, and because
the matrix-vector product to compute the output pre-activations for the
underlying classifier is restricted to a potentially different set of
candidates for each example.

Tree recall is 81.3\% and 50.4\% on the training and test set
respectively: by comparison, the recall of the 4000 most frequent training
set labels is 31.6\% and 29.3\% on the training and test set respectively.
The test set deviation of the recall is somewhat disappointing, but
nonetheless overall accuracy is superior for the current method.

\cite{choromanska2014logarithmic} report 6.57\% test accuracy on this
data set.

\subsection{LSHTC} 

The Large Scale Hierarchical Text Classification Challenge
was a public competition involving multilabel classification of documents
into approximately 300,000 categories \citep{LSHTC4}. The training
and (unlabeled) test files are available from the Kaggle platform.
The features are bag of words representations of each document.

We used a tree with maximum depth 14, maximum classifiers per leaf of
4000, and acceptable training tree recall of 99.9\%.  Preprocessing 
is identical to ODP.

The node indicator variable is augmented to the feature representation but not
interacted, i.e., each node learns a distinct bias for each (candidate)
class, but otherwise shares all parameters.  Training and test set
recall for the tree are 86\% and 72\% respectively; by comparison, the
recall of the 4000 most frequent training set labels is 33.8\% and 33.7\%
on the training and test set respectively.

\begin{table}[h]
\centering
\caption{LSHTC results.  Averages are example-averages across the training set.}
\begin{tabular}{|c|c|c|} \hline 
Model                                          & Rank-800 ILR        & Tree + ILR  \\ \hline
{\centering \vspace{1pt} Avg Depth}            & n/a                 & 14 \\ \hline
{\centering \vspace{1pt} Avg Leaf Classifiers} & n/a                 & 3998 \\ \hline
{\centering \vspace{1pt} Test Tree Recall}     & n/a                 & 72\% \\ \hline
{\centering \vspace{1pt} Test Precision@1}     & 53.4\%              & 54.0\% \\ \hline
{\centering \vspace{1pt} Inference Speed}      & 60 ex/s             & 1058 ex/s \\ \hline
\end{tabular}
\label{tab:lshtcres1}
\end{table}


We compare with \cite{mineiro2015fast} in Table~\ref{tab:lshtcres1},
using the same train-test split as that publication.  The baseline is a
rank-800 per-class approximate kernel logistic regression trained via
randomized techniques. This model is equivalent to a 2 hidden layer
neural network, where the first hidden layer of 800 units has linear
activation; the second hidden layer of 4000 units has cosine activation,
i.e., random Fourier features \citep{rahimi2007random}; and the final
layer of 325K units minimizes cross-entropy loss.
Computationally, the baseline is far slower than the current technique.
Statistically, performance is similar.

\begin{table}[h]
\centering
\caption{More LSHTC results.  The train-test split is different than that of table~\ref{tab:lshtcres1}.  Starred timings are as reported by other authors using presumably different hardware.}
\begin{tabular}{|c|c|c|c|c|} \hline 
\multirow{2}{*}{Model} & \multicolumn{3}{|c|}{Precision at} & \parbox{38pt}{\centering \vspace{1ex} Inference \\ Speed \\ (ex/s) \vspace{-1ex}}\\ \cline{2-4}
& 1 & 3 & 5 & \\ \hline
FastXML & 49.35\% & 32.69\% & 24.03\% & 2000${}^*$ \\
X1 & 55.57\% & 33.84\% & 24.07\% & 125${}^*$ \\ \hline
Tree + ILR & 53.0\% & 33.9\% & 24.8\% & 1058 \\ 
$k=6000$ & 53.7\% & 34.5\% & 25.3\% & 688 \\
$k=12000$ & 54.3\% & 34.9\% & 25.6\% & 370  \\
\hline
\end{tabular}
\label{tab:lshtcres2}
\end{table}

We also compare with published results for FastXML
\citep{prabhu2014fastxml} and X1 \citep{bhatia2015locally} in
Table~\ref{tab:lshtcres2} using the same train-test split as those
publications, which is a different train-test split than that utilized
in Table~\ref{tab:lshtcres1}.  Reported metrics for FastXML and
X1 are for ensembles of these models, with 50 and 15 elements in each
ensemble respectively.\footnote{It is unclear if timings are
for a single model or the ensemble.}  For our model, we show several
results corresponding to different numbers of classifiers at each leaf:
these correspond to different points on a Pareto frontier trading
off accuracy with inference time.  FastXML does not compare
favorably, but there is no clear preference between X1 and our model: X1 has
superior precision at 1, and our model has superior precision at 3 and 5.

\section{Discussion}

The technique presented in this paper attempts to learn a particular type
of output structure, namely regions of feature space for which a small
number of labels tend to occur.  This structure is clearly useful for the
particular inference strategy articulated in Algorithm~\ref{alg:cset}, but
has other potential applications.  For example, during exploratory data
analysis, it might be useful to identify labels that are cooccur in the same regions of feature space.

The Twitter experiment suggests the most frequent classes are always part
of the candidate set.  One future direction of research is to investigate
whether head classes should treated separately from tail classes,
e.g., only use the tree to identify candidates amongst tail classes.
This might mitigate the need to have the total leaf node candidate slots
greatly exceed the number of classes.

One natural question is why both a filter and a classifier are required:
an alternate strategy would be to repeatedly filter until a simple model
is invoked, e.g., a constant predictor, which for multiclass would be
a single class and for multilabel with ranking loss would be a fixed
ordering over the classes.  Our initial (limited) experimentation did
not yield promising results in this direction.  We speculate two possible
issues: the difference in decision surfaces expressible by one-versus-all
compared to conjunctions of hyperplanes, and the high sample complexity
of learning deep trees due to data decimation.

We obtain good results for text problems, where linear predictors have
good performance.  We also expect to do well on problems where (primal
approximations of) kernel machines have good performance.  However, we do
not anticipate good performance for direct application of this technique
to problems where a feature representation must be learned, e.g., image
classification.  For these problems it is implausible that hyperplanes
in the original feature space will produce a high recall label filter.
Instead, one possibility is to use the feature representation from a
smaller (number of labels) version of the problem to drive the tree
construction for the original problem, e.g., use of an internal layer
of a pretrained convolutional network as a feature map.  Another is
to replace the eigenvalue strategy at each node with a procedure that
finds a one-dimensional nonlinear function of the data which is highly
correlated with the labels, e.g., deep CCA \citep{andrew2013deep}.


\bibliography{spectral_tree}

\begin{thebibliography}{24}
\providecommand{\natexlab}[1]{#1}
\providecommand{\url}[1]{\texttt{#1}}
\expandafter\ifx\csname urlstyle\endcsname\relax
  \providecommand{\doi}[1]{doi: #1}\else
  \providecommand{\doi}{doi: \begingroup \urlstyle{rm}\Url}\fi

\bibitem[ODP()]{ODP}


\bibitem[Andrew et~al.(2013)Andrew, Arora, Bilmes, and Livescu]{andrew2013deep}
Andrew, Galen, Arora, Raman, Bilmes, Jeff, and Livescu, Karen.
\newblock Deep canonical correlation analysis.
\newblock In \emph{Proceedings of the 30th International Conference on Machine
  Learning}, pp.\  1247--1255, 2013.

\bibitem[Bengio et~al.(2010)Bengio, Weston, and Grangier]{bengio2010label}
Bengio, Samy, Weston, Jason, and Grangier, David.
\newblock Label embedding trees for large multi-class tasks.
\newblock In \emph{Advances in Neural Information Processing Systems}, pp.\
  163--171, 2010.

\bibitem[Bennett \& Nguyen(2009)Bennett and Nguyen]{bennett2009refined}
Bennett, Paul~N and Nguyen, Nam.
\newblock Refined experts: improving classification in large taxonomies.
\newblock In \emph{Proceedings of the 32nd international ACM SIGIR conference
  on Research and development in information retrieval}, pp.\  11--18. ACM,
  2009.

\bibitem[Bhatia et~al.(2015)Bhatia, Jain, Kar, Varma, and
  Jain]{bhatia2015locally}
Bhatia, Kush, Jain, Himanshu, Kar, Purushottam, Varma, Manik, and Jain,
  Prateek.
\newblock Sparse local embeddings for extreme multi-label classification.
\newblock In \emph{Advances in Neural Information Processing Systems}, pp.\
  730--738, 2015.

\bibitem[Caruana(1997)]{caruana1997multitask}
Caruana, Rich.
\newblock Multitask learning.
\newblock \emph{Machine learning}, 28\penalty0 (1):\penalty0 41--75, 1997.

\bibitem[Chang et~al.(2015)Chang, Krishnamurthy, Agarwal, Daume, and
  Langford]{chang2015learning}
Chang, Kai-wei, Krishnamurthy, Akshay, Agarwal, Alekh, Daume, Hal, and
  Langford, John.
\newblock Learning to search better than your teacher.
\newblock In \emph{Proceedings of the 32nd International Conference on Machine
  Learning (ICML-15)}, pp.\  2058--2066, 2015.

\bibitem[Chen et~al.(2011)Chen, Zhou, and Ye]{chen2011integrating}
Chen, Jianhui, Zhou, Jiayu, and Ye, Jieping.
\newblock Integrating low-rank and group-sparse structures for robust
  multi-task learning.
\newblock In \emph{Proceedings of the 17th ACM SIGKDD international conference
  on Knowledge discovery and data mining}, pp.\  42--50. ACM, 2011.

\bibitem[Choromanska \& Langford(2015)Choromanska and
  Langford]{choromanska2014logarithmic}
Choromanska, Anna~E and Langford, John.
\newblock Logarithmic time online multiclass prediction.
\newblock In \emph{Advances in Neural Information Processing Systems}, pp.\
  55--63, 2015.

\bibitem[Daum{\'e}~III et~al.(2014)Daum{\'e}~III, Langford, and
  Ross]{daume2014efficient}
Daum{\'e}~III, Hal, Langford, John, and Ross, Stephane.
\newblock Efficient programmable learning to search.
\newblock \emph{arXiv preprint arXiv:1406.1837}, 2014.

\bibitem[Evgeniou \& Pontil(2007)Evgeniou and Pontil]{evgeniou2007multi}
Evgeniou, A and Pontil, Massimiliano.
\newblock Multi-task feature learning.
\newblock \emph{Advances in neural information processing systems},
  19:\penalty0 41, 2007.

\bibitem[Geusebroek et~al.(2005)Geusebroek, Burghouts, and
  Smeulders]{geusebroek2005amsterdam}
Geusebroek, Jan-Mark, Burghouts, Gertjan~J, and Smeulders, Arnold~WM.
\newblock The {A}msterdam library of object images.
\newblock \emph{International Journal of Computer Vision}, 61\penalty0
  (1):\penalty0 103--112, 2005.

\bibitem[Halko et~al.(2011)Halko, Martinsson, and Tropp]{halko2011finding}
Halko, Nathan, Martinsson, Per-Gunnar, and Tropp, Joel~A.
\newblock Finding structure with randomness: Probabilistic algorithms for
  constructing approximate matrix decompositions.
\newblock \emph{SIAM review}, 53\penalty0 (2):\penalty0 217--288, 2011.

\bibitem[Mineiro \& Karampatziakis(2015)Mineiro and
  Karampatziakis]{mineiro2015fast}
Mineiro, Paul and Karampatziakis, Nikos.
\newblock Fast label embeddings via randomized linear algebra.
\newblock In \emph{Machine Learning and Knowledge Discovery in Databases}, pp.\
   37--51. Springer, 2015.

\bibitem[Mnih \& Hinton(2009)Mnih and Hinton]{mnih2009scalable}
Mnih, Andriy and Hinton, Geoffrey~E.
\newblock A scalable hierarchical distributed language model.
\newblock In \emph{Advances in neural information processing systems}, pp.\
  1081--1088, 2009.

\bibitem[Morin \& Bengio(2005)Morin and Bengio]{morin2005hierarchical}
Morin, Frederic and Bengio, Yoshua.
\newblock Hierarchical probabilistic neural network language model.
\newblock In \emph{Proceedings of the international workshop on artificial
  intelligence and statistics}, pp.\  246--252. Citeseer, 2005.

\bibitem[Partalas et~al.(2015)Partalas, Kosmopoulos, Baskiotis, Artieres,
  Paliouras, Gaussier, Androutsopoulos, Amini, and Galinari]{LSHTC4}
Partalas, Ioannis, Kosmopoulos, Aris, Baskiotis, Nicolas, Artieres, Thierry,
  Paliouras, George, Gaussier, Eric, Androutsopoulos, Ion, Amini, Massih-Reza,
  and Galinari, Patrick.
\newblock {LSHTC}: A benchmark for large-scale text classification.
\newblock \emph{arXiv preprint arXiv:1503.08581}, 2015.

\bibitem[Prabhu \& Varma(2014)Prabhu and Varma]{prabhu2014fastxml}
Prabhu, Yashoteja and Varma, Manik.
\newblock Fastxml: A fast, accurate and stable tree-classifier for extreme
  multi-label learning.
\newblock In \emph{Proceedings of the 20th ACM SIGKDD international conference
  on Knowledge discovery and data mining}, pp.\  263--272. ACM, 2014.

\bibitem[Rahimi \& Recht(2007)Rahimi and Recht]{rahimi2007random}
Rahimi, Ali and Recht, Benjamin.
\newblock Random features for large-scale kernel machines.
\newblock In \emph{Advances in neural information processing systems}, pp.\
  1177--1184, 2007.

\bibitem[Rifkin \& Klautau(2004)Rifkin and Klautau]{rifkin2004defense}
Rifkin, Ryan and Klautau, Aldebaro.
\newblock In defense of one-vs-all classification.
\newblock \emph{The Journal of Machine Learning Research}, 5:\penalty0
  101--141, 2004.

\bibitem[Robertson et~al.(1999)Robertson, Walker, Beaulieu, and
  Willett]{robertson1999okapi}
Robertson, Stephen~E, Walker, Steve, Beaulieu, Micheline, and Willett, Peter.
\newblock Okapi at trec-7: automatic ad hoc, filtering, vlc and interactive
  track.
\newblock \emph{Nist Special Publication SP}, pp.\  253--264, 1999.

\bibitem[Weinberger et~al.(2009)Weinberger, Dasgupta, Langford, Smola, and
  Attenberg]{weinberger2009feature}
Weinberger, Kilian, Dasgupta, Anirban, Langford, John, Smola, Alex, and
  Attenberg, Josh.
\newblock Feature hashing for large scale multitask learning.
\newblock In \emph{Proceedings of the 26th Annual International Conference on
  Machine Learning}, pp.\  1113--1120. ACM, 2009.

\bibitem[Weston et~al.(2013)Weston, Makadia, and Yee]{weston2013label}
Weston, Jason, Makadia, Ameesh, and Yee, Hector.
\newblock Label partitioning for sublinear ranking.
\newblock In \emph{Proceedings of the 30th International Conference on Machine
  Learning (ICML-13)}, pp.\  181--189, 2013.

\bibitem[Yildiz \& Alpaydin(2005)Yildiz and Alpaydin]{yildiz2005linear}
Yildiz, Olcay~Taner and Alpaydin, Ethem.
\newblock Linear discriminant trees.
\newblock \emph{International Journal of Pattern Recognition and Artificial
  Intelligence}, 19\penalty0 (03):\penalty0 323--353, 2005.

\end{thebibliography}

\bibliographystyle{icml2016}

\end{document}